%% file: root.tex
\documentclass[letterpaper, 10 pt, conference]{ieeeconf}  % Comment this line out if you need a4paper

\IEEEoverridecommandlockouts                              % This command is only needed if 
\usepackage{amsmath,amsfonts}
\usepackage{amssymb}

\usepackage{algorithm}
\usepackage[noend]{algorithmic}
\usepackage{color}
\usepackage{caption}

\usepackage{array}
\usepackage{textcomp}
\usepackage{stfloats}

\usepackage{verbatim}
\usepackage{graphicx}
\usepackage{wasysym}
\let\labelindent\relax
\usepackage{enumitem}
\usepackage{threeparttable}
\usepackage{makecell}

\usepackage{cite}
\usepackage[utf8]{inputenc} % allow utf-8 input
\usepackage[T1]{fontenc}    % use 8-bit T1 fonts
\usepackage{hyperref}       % hyperlinks
\usepackage{url}            % simple URL typesetting
\usepackage{booktabs}       % professional-quality tables
\usepackage{amsfonts}       % blackboard math symbols
\usepackage{nicefrac}       % compact symbols for 1/2, etc.
\usepackage{microtype}      % microtypography
\usepackage{lipsum}
\usepackage{fancyhdr}       % header
\usepackage{graphicx}       % graphics
\usepackage{wrapfig}
\usepackage{subfigure}
\usepackage{multirow}

\title{\LARGE \bf
REBot: Reflexive Evasion Robot for Instantaneous \\ Dynamic Obstacle Avoidance
}

\author{Zihao Xu$^{1}$, Ce Hao$^{1}$, Chunzheng Wang$^{2}$, Kuankuan Sima$^{2}$, Fan Shi$^{2}$, and Jin Song Dong$^{1}$
\thanks{$^1$ School of Computing, National University of Singapore, Singapore.}
\thanks{$^2$ Department of Electrical and Computer Engineering, National University of Singapore, Singapore.}
}

\begin{document}

\maketitle
\thispagestyle{empty}
\pagestyle{empty}

%%%%%%%%%%%%%%%%%%%%%%%%%%%%%%%%%%%%%%%%%%%%%%%%%%%%%%%%%%%%%%%%%%%%%%%%%%%%%%%%

\input{sections/0_Abstract}
\input{sections/1_Introduction}

\input{sections/2_Related_works}
\input{sections/3_Preliminary}
\input{sections/4_Method}
\input{sections/5_Experiment}
\input{sections/6_real_robot}

\input{sections/7_conclusion}

%%%%%%%%%%%%%%%%%%%%%%%%%%%%%%%%%%%%%%%%%%%%%%%%%%%%%%%%%%%%%%%%%%%%%%%%%%%%%%%%
\bibliographystyle{IEEEtran}
\bibliography{references}

\clearpage
\appendices
\input{sections/11_appendix}

\end{document}

%% file: sections/0_Abstract.tex
\begin{abstract}
Dynamic obstacle avoidance (DOA) is critical for quadrupedal robots operating in environments with moving obstacles or humans. Existing approaches typically rely on navigation-based trajectory replanning, which assumes sufficient reaction time and leading to fails when obstacles approach rapidly. In such scenarios, quadrupedal robots require reflexive evasion capabilities to perform instantaneous, low-latency maneuvers. This paper introduces Reflexive Evasion Robot (REBot), a control framework that enables quadrupedal robots to achieve real-time reflexive obstacle avoidance. REBot integrates an avoidance policy and a recovery policy within a finite-state machine. With carefully designed learning curricula and by incorporating regularization and adaptive rewards, REBot achieves robust evasion and rapid stabilization in instantaneous DOA tasks. We validate REBot through extensive simulations and real-world experiments, demonstrating notable improvements in avoidance success rates, energy efficiency, and robustness to fast-moving obstacles. Videos and appendix are available on \href{https://rebot-2025.github.io/}{https://rebot-2025.github.io/}.
\end{abstract}

%% file: sections/1_Introduction.tex
\begin{figure*}[t]
    \centering
    \vspace{5mm}
    \includegraphics[width=0.80\textwidth]{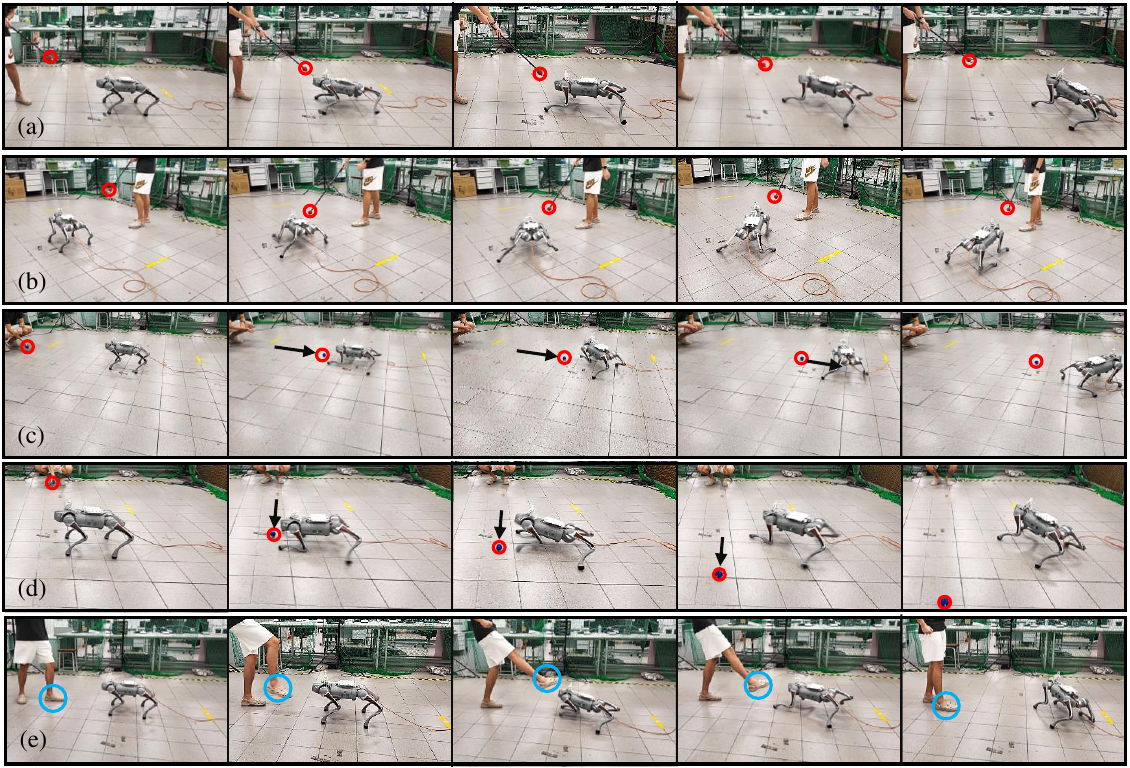}
    \begin{small}
        \caption{The Reflexive Evasion Robot (\textbf{REBot}) system achieves instantaneous dynamic obstacle avoidance. When the fast-moving obstacles approach the quadrupedal robots (reaction time$<$1.5s), REBot switches to the avoidance policy and performs reflexive evasion maneuvers. In (a) and (b), the robot is poked on the frontal and dorsal sides using a stick; in (c) and (d), a ball is launched toward the robot from both frontal and lateral directions; in (e), to evaluate robustness, the quadrupedal robot was subjected to intentional kicks from multiple directions. Experimental results demonstrate that the REBot system successfully controlled the quadrupedal robot to avoid all obstacles.}
        \label{Fig: teaser}
    \end{small}
\vspace{-4mm}
\end{figure*}

\section{Introduction} \label{Sec: Introduction}

Ensuring the safety of a robot is essential during task execution~\cite{shi2024rethinking}. For instance, a mobile robot must not only perform its primary tasks but also perceive surrounding obstacles and take appropriate evasive actions~\cite{tao2024mobile,sun2025sparkmodularbenchmarkhumanoid,falanga2020dynamic}. When encountering slow-moving obstacles (reaction time $>$2s), the robot typically has sufficient time to stop its current actions and replan a new trajectory using a decision-making model to avoid collisions~\cite{hoeller2021learning,dudzik2020robust}. This type of behavior is commonly referred to as dynamic obstacle avoidance (DOA) via navigation-based trajectory replanning~\cite{yang2021learning}.
For legged robots, such as quadrupedal robots, DOA involves both high-level navigation decision-making and low-level locomotion control~\cite{huang2023creating}. For example, in the Agile but Safe (ABS) framework~\cite{ABS}, a quadrupedal robot encountering a quasi-static obstacle during high-speed locomotion can reduce its speed and replan a navigation trajectory to safely bypass the obstacle.

However, when obstacles approach at high speeds, the robot is left with extremely limited reaction time ($<$1.5s), necessitating immediate evasive maneuvers~\cite{lu2024fapp}. Due to limitations in mechanical structure and motor power, the robot often fails to generate sufficient velocity within the available time to accurately track a replanned navigation trajectory~\cite{li2023robust}.
To achieve instantaneous DOA, we draw inspiration from the spinal reflex systems of vertebrates. Unlike decision-making processes governed by the brain, spinal reflexes enable rapid, localized decisions through neural circuits in the spinal cord, allowing animals to execute unconventional evasive actions instantaneously~\cite{umeda2024future}. For example, an antelope might execute a sudden backward leap to evade an ambush from an underwater crocodile while drinking, relying entirely on reflexive evasion.

In this paper, we propose the Reflexive Evasion Robot (REBot) system for instantaneous dynamic obstacle avoidance. Using the quadrupedal robot Unitree Go2 as an example platform~\cite{liu2024deployment,xiao2024egocentric}, REBot demonstrates real-time evasion of high-speed obstacles with a reaction time of less than 1.5 seconds.
The REBot system is structured as a finite-state machine with three behavioral stages. During the normal stage, the robot performs its primary functional tasks. When an approaching obstacle is detected, REBot transitions to the avoidance stage, executing reflexive evasion maneuvers.
During evasion, reinforcement learning (RL) policy~\cite{schulman2017proximal,gurram2025reinforcement,xu2024dexterous} trained by PPO enables rapid avoidance while preserving the robot’s safety, balance, and energy efficiency. After an evasive maneuver, the robot may become unstable. REBot then enters the recovery stage, during which a policy stabilizes the robot and restores normal function.

We trained the REBot system for quadrupedal robots in Isaac Gym simulator~\cite{makoviychuk2021isaac}, evaluated its performance, and deployed it on a real robot for demonstration.
REBot achieved the highest avoidance and recovery success rates in both static and dynamic obstacle scenarios, while reducing maximum joint power and avoidance distance.
We observed that the robot's reflexive evasion performance varied with obstacle direction and speed; it performed best when avoiding frontal obstacles due to its structural advantages in backward maneuvers.
Ablation studies confirmed that the recovery policy, curriculum learning, and adaptive reward design significantly improved avoidance success rates.
Finally, real-world experiments (Fig.~\ref{Fig: teaser}) validated REBot’s capability for real-time, instantaneous dynamic obstacle avoidance, offering insights into robot safety system design.

In summary, the contributions of this paper are as follows: 
\vspace{-2mm}
\begin{itemize}[leftmargin=15pt] 
    \item We formally identify and formulate the reflexive evasion problem for dynamic obstacle avoidance under constrained reaction time in quadrupedal robots. 
    \item We design the REBot system as a finite-state machine integrating avoidance and recovery policies to achieve robust, real-time reflexive evasion. 
    \item We conduct comprehensive simulations and real-world experiments with thorough analysis to validate REBot’s effectiveness across various obstacle scenarios. 
\end{itemize}

%% file: sections/2_Related_works.tex
\section{Related Works} \label{Sec: Related Works}

Quadrupedal robots have achieved significant progress in recent years~\cite{tranzatto2022cerberus,sun2020path}, enabled by both model-based and learning-based control frameworks. Model-based approaches such as zero-moment point (ZMP) planning~\cite{meng2023trot,chen2022adaptive}, centroidal dynamics optimization~\cite{zhou2022momentum,chi2022linearization}, and Model Predictive Control (MPC)~\cite{gangapurwala2022rloc,hwangbo2019learning} enable precise trajectory tracking and support dynamic maneuvers under unstructured environments. In parallel, RL has empowered quadrupeds with a wide range of locomotion skills~\cite{gurram2025reinforcement}, from diverse gaits~\cite{wu2023learning} to robust terrain traversal~\cite{luo2024moral} and morphology adaptation~\cite{zhang2024synloco}. RL has also been applied to static obstacle avoidance~\cite{ABS}, where methods like ABS combine learned control with navigation planning to produce collision-free paths. However, such methods rely heavily on trajectory replanning and sufficient reaction time, which limits their applicability under fast or unpredictable conditions.

Dynamic obstacle avoidance has been extensively studied across UAVs~\cite{azzam2023learning,falanga2020dynamic,lu2024fapp,Fan_2025}, mobile robots~\cite{wang2021review,tao2024mobile}, humanoids~\cite{kim2024armor,sun2025sparkmodularbenchmarkhumanoid}, and manipulators~\cite{song2024mpc,zhang2024catch}. UAVs leverage high-frequency planning and reactive control~\cite{falanga2020dynamic,lu2024fapp}, while Fan et al.~\cite{Fan_2025} proposed a deep RL-based policy that enables fast, map-free evasive maneuvers against dynamic threats, even at high relative speeds. In mobile platforms, RL frameworks with motion constraints~\cite{tao2024mobile} improve navigation in crowded environments. For humanoids and manipulators, full-body coordination has been exploited to intercept or avoid moving objects~\cite{zhang2024catch,sun2025sparkmodularbenchmarkhumanoid}. These advances demonstrate diverse platform-specific strategies for dynamic avoidance, but few translate directly to legged locomotion.

Despite these advances, quadrupedal robots currently lack general-purpose frameworks for dynamic obstacle avoidance under constrained reaction time. Existing approaches often rely on predefined motion primitives such as sidestepping, without the reflexive strategies seen in other platforms. Bridging this gap remains an open challenge and is critical for deploying legged robots in fast-changing environments.

%% file: sections/3_Preliminary.tex
\section{Preliminary} \label{Sec: Preliminary}

\begin{figure}[t]
    \centering
    \vspace{5mm}
    \includegraphics[width=0.85\columnwidth]{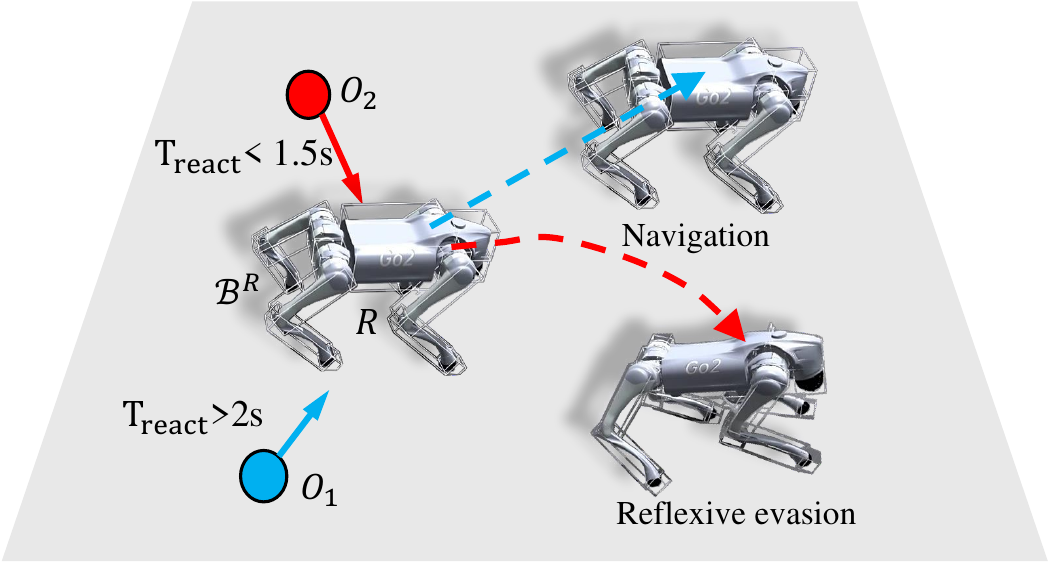}
    \caption{Robot dynamic obstacle avoidance based on different reaction time. Red line: reflexive evasion under shorter reaction time; blue line: navigation\_based avoidance under longer reaction time.}
    \label{Fig: prelim}
\vspace{-5mm}
\end{figure}

\textbf{Problem formulation}. 
Fig.~\ref{Fig: prelim} shows that the dynamic obstacle avoidance (DOA) system has two entities: dynamic obstacles and quadruped robots. 
The dynamic obstacles $O$ are modeled as a rigid sphere with states $(r^O, p^O_t, v^O_t)$ of radius, position, velocity and acceleration in 3D space. 
The quadruped robot $R$ is a high-dimensional articulated system with a robot base and four independently actuated legs. The states $s^R_t$ consist of base position $p^R_t \in \mathbb{R}^3$, linear velocity $v^R_t \in \mathbb{R}^3$, angular velocity $\omega^R_t \in \mathbb{R}^3$, joint position $q^R_t \in \mathbb{R}^{12}$, joint velocity $\dot{q}^R_t \in \mathbb{R}^{12}$, joint torque $\tau^R_t \in \mathbb{R}^{12}$, contact force $f^R_t \in \mathbb{R}^{4 \times 3}$ and orientation angle $\theta^R_t \in \mathbb{R}^3$. The robot is driven by servo motors on the joints to move on the ground via action $a^R_t$, where $a^R_t \in \mathbb{R}^{12}$ denotes the joint target angles. 
In this work, we utilize the Unitree Go2 robot to conduct experiments.
We define a successful dynamic obstacle avoidance (DOA) as the robot maintaining collision-free motion throughout the task duration. A collision is considered to occur if the signed distance function (SDF) from the obstacle center $p^O_t$ to the robot's oriented bounding box (OBB), denoted as $\mathcal{B}^R$, is smaller than the obstacle's radius $r^O$; that is, if $d(p^O_t, \mathcal{B}^R) < r^O$.

\textbf{Dynamic obstacle Avoidance Strategies}. 
The robot adopts a control system with observation $(p^R_t, \omega^R_t, q^R_t, \dot{q}^R_t, \tau^R_t, f^R_t, p^O_t, v^O_t, r^O)$ and action $(a^R_t)$, enabling the robot to avoid approaching obstacles while maintaining balance. 
When faced with static or slow-moving obstacles, the robot can temporarily stop and replan its trajectory, a behavior classified as \textbf{navigation avoidance}~\cite{ABS}.
However, when the obstacles approach instantaneously, the robot must react immediately. We categorize such behavior as \textbf{reflexive evasion} (Fig.~\ref{Fig: prelim}). The reaction time $T_{\text{react}}$ determines the reaction of navigation or reflex, and we distinguish these two behaviors by the maximum joint power of the robot. In the following sections, we design the REBot system to achieve the reflexive DOA and analyze the behaviors in both simulation and real-world experiments.

%% file: sections/4_Method.tex
\section{Method} \label{Sec: Method}

\begin{figure*}[t]
    \centering
    \vspace{5mm}
    \includegraphics[width=0.80\textwidth]{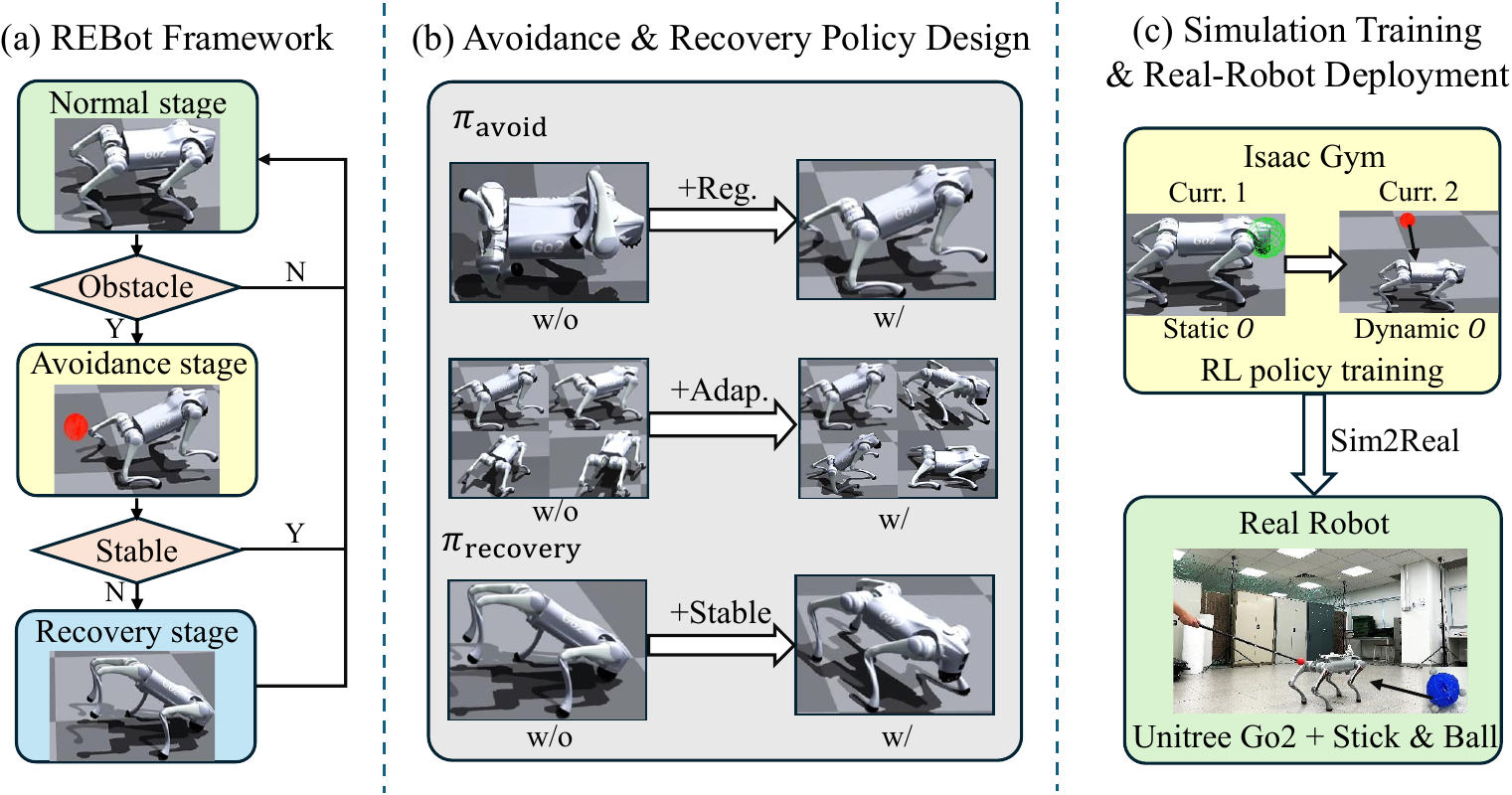}
    \caption{\textbf{(a) REBot Framework:} A finite-state machine (FSM) governs transitions between the Normal, Avoidance, and Recovery stages. The quadrupedal robot performs reflexive evasion to avoid obstacles (red balls).
    \textbf{(b) Policy Design:} The avoidance policy is trained not only for successful obstacle avoidance but also incorporates regulation rewards for state stabilization and adaptive rewards to encourage diverse evasive behaviors.
    \textbf{(c) Training and Deployment:} REBot is trained in Isaac Gym using a two-stage curriculum from static to dynamic obstacles, and deployed on a real Unitree Go2 robot.}
    \label{Fig: Pipeline}
\vspace{-5mm}
\end{figure*}

In this section, we design the reflexive evasion robot (\textbf{REBot}) system to achieve DOA in Fig.~\ref{Fig: Pipeline}(a). As shown in Fig.~\ref{Fig: Pipeline}, the REBot system consists of three behavioral stages organized as a finite state machine (FSM). In the following sections, we introduce the FSM stages and transition criteria (Sec. \ref{Subsec: stage transition}), the training strategies of the avoidance policy (Sec.~\ref{Subsec: Avoidance Policy Training}) and the recovery policy (Sec.~\ref{Subsec: Recovery Policy Training}). Finally, we delineate the training and deployment of the REBot system in simulation and on real quadruped robots, respectively (Sec.~\ref{Subsec: real robot deploy}).

\subsection{REBot Stages and Transition Criteria} \label{Subsec: stage transition}

The robot initially stays in the \textbf{normal stage} with functional behaviors such as standing, walking, or trotting. During this stage, the robot continuously observes the environment and potential obstacles. In this work, we define the normal behavior as standing still while maintaining balance via the PD controller. 
When the obstacle is approaching the robot, (i.e., $ \langle v^O_t,\, p^R_t - p^O_t \rangle > 0 $), REBot switches to the \textbf{avoidance stage} to execute reflexive evasion, in which the avoidance policy performs reactive maneuvers under constrained reaction time. 

However, severe evasion may cause the robots' instability. Therefore, the REBot correspondingly switches to the \textbf{recovery stage}, where a recovery policy drives the robot back to normal functions. REBot judges the instability with three criteria: 
(i) body orientation exceeds a safe range \( \|\theta^R_t\| > \theta^R_{\text{th}} \); (ii) joint velocity surpasses a stability limit \( \|\dot{q}^{R}_t\| > \dot{q}^{R}_{\text{th}} \); (iii) base height drops below a threshold value \( h^R_t < h^R_{\text{th}} \). Here, \( \theta^R_t  \), \( \dot{q}^{R}_t \), and \( h^R_t \in \mathbb{R} \) denote the robot's orientation, joint velocity, and base height, respectively. The corresponding thresholds are \( \theta^R_{\text{th}} \), \( \dot{q}^{R}_{\text{th}} \), and \( h^R_{\text{th}} \).

\subsection{Avoidance Policy} \label{Subsec: Avoidance Policy Training}

The avoidance policy is trained via RL to achieve reflexive evasion under constrained reaction time. The objective is to avoid collisions while maintaining postural stability and minimizing energy consumption. 
The reward function consists of three parts: $r = r_{\text{avoidance}} + r_{\text{regularization}} + r_{\text{adaptive}}$.

\textbf{Avoidance reward} is composed of $r_{\text{avoidance}} = r_{\text{distance}} + r_{\text{collision}}$. The distance reward is defined as $r_{\text{distance}} = -\exp(-(d(p^O_t, \mathcal{B}^R) - r^O))$ to encourage the robot to maintain a safe distance from the moving obstacle throughout the entire task. The collision penalty is defined as $r_{\text{collision}} = \mathbf{1}(c = 0) - \mathbf{1}(c = 1)$ to penalize any contact event, where $c \in \{0,1\}$ denotes the collision. 
We adopt a two-stage curriculum to improve policy training efficiency. In the first stage, a static obstacle appears instantaneously in a random location near the robot at a predefined activation time. In the second stage, the obstacle follows a directed trajectory toward the robot at varying speeds, simulating realistic threats. 

\textbf{Regularization reward} consists of three terms $r_{\text{regularization}} = r_{\text{walk}} + r_{\text{energy}} + r_{\text{contact}}$. It is designed to ensure the learned evasive maneuvers remain both stable and natural (Fig.~\ref{Fig: Pipeline}(b)). 
To promote natural and coordinated motion, we encourage symmetric limb phasing consistent with a trot gait. The term $r_{\text{walk}} = \frac{1}{2} \left( \mathbf{1}(c_{\text{FL}} = c_{\text{RR}}) + \mathbf{1}(c_{\text{FR}} = c_{\text{RL}}) \right)$ rewards synchronized contact patterns between diagonal leg pairs, where $c_{i,j}$ denotes different contact leg.
We penalize the product of joint torque and joint velocity across all actuated degrees of freedom to reduce excessive power consumption through $r_{\text{energy}} = -\sum_{i} | \tau_t^{R,i} \cdot \dot{q}_t^{R,i} |$, where $\tau_t^{R,i}$ and $\dot{q}_t^{R,i}$ represent the torque and angular velocity of each joint respectively. 
We also penalize the temporal fluctuation of vertical foot contact forces to reduce instability via $r_{\text{contact}} = - \sum_{i} \left( f_{t}^{R,i,z} - f_{t-1}^{R,i,z} \right)^2$, where $f_{t}^{R,i,z}$ denotes vertical foot contact force of each leg.

\textbf{Adaptive reward} is designed as $r_{\text{adaptive}} = r_{\text{diversity}} + r_{\text{threat}} + r_{\text{direction}}$ to encourage motion diversity, speed modulation, and direction efficiency~\cite{eysenbach2018diversity}. 
RL policy tends to converge toward a single locally optimal behavior (Fig.~\ref{Fig: Pipeline}(b)). In this task, it manifests as the robot repeatedly using a fixed evasion gait regardless of obstacle state. We defined a diversity reward to encourage the policy to appropriate behaviors $r_{\text{diversity}} = \text{Var}_{s^R \sim \mathcal{D}_{s^R}} \left[ \pi(a^R_t|s^R_t) \right]$. 
We define the threat level of the obstacle through the reaction time and the robot learns to adapt its speed in response to the levels of perceived threat as
$r_{\text{threat}} = -\| v_t^{\text{R}} - v_t^{\text{R,safe}} \|, \quad v_t^{\text{R,safe}} = v_t^{\text{R,cmd}} + \lambda \exp(-\eta T_{\text{reaction}})$,
where $v_t^{\text{R,cmd}}$ denotes the command velocity, $\lambda$ and $\eta$ denote the hyperparameters.
To discourage evasive movements that deviate unnecessarily from the ideal escape direction, we penalize wrong the robot movement direction via $r_{\text{direction}} = -\langle v_t^{R},\,  p^O_t - p_t^{R} \rangle$. 

\subsection{Recovery Policy} \label{Subsec: Recovery Policy Training}

The recovery policy ensures a smooth transition from the avoidance stage back to the normal stage, allowing the robot to regain balance (Fig.~\ref{Fig: Pipeline}(b)). 
Therefore, the reward function $r = r_{\text{orientation}} + r_{\text{stable}} + r_{\text{position}} + r_{\text{additional}}$ is designed corresponding to the instability criteria. The orientation reward defined as $r_{\text{orientation}} = - \sum_i (\theta_t^{R,i} - \theta_0^{R,i})^2$ penalizes excessive tilt of the robot, where $\theta_0^{R,i} \in \mathbb{R}$ denotes the default orientation angles. The stable reward $r_{\text{stable}} = \sum_i \exp(-|\dot{q}_t^{R,i}|)$ encourages low joint velocities via exponential decay. And the position term $r_{\text{position}} = -\| p_t^R - p_{0}^R \|^2$ penalizes too slow base height, where $p_{0}^R$ denotes the default position.
The additional reward term $r_\text{additional}$ includes penalties on large joint torque and action discontinuities. These components are designed to reduce abrupt joint movements and encourage smoother transitions during recovery.

\subsection{Training in Simulation and Real-Robot Deployment} \label{Subsec: real robot deploy}

We implement the REBot system on the Unitree Go2 quadrupedal robot (Fig.~\ref{Fig: Pipeline}(c)). We train the avoidance and recovery policies in the Isaac gym simulator~\cite{rudin2022learning} with PPO algorithm ~\cite{han2024learning,zhao2024zsl,mock2023comparison}. 
Specifically, the avoidance policy is trained in two curricula. First, a stationary obstacle is randomly placed around the robot and activates after a delay; Go2 must react within the available response time. Second, a moving obstacle approaches with fixed velocity from a random direction, requiring real-time evasion. 
Both curricula randomize obstacle parameters to prevent overfitting and encourage generalization across planning-based and reflexive behaviors.
Then we deploy the REBot system to the real Unitree Go2 robot. A motion capture system was used to provide real-time position data for both the robot and the dynamic obstacle. To emulate dynamic obstacles, we used a rigid rod with a lightweight ball attached to its tip, serving as a physical proxy for an incoming object. The obstacle's position was continuously tracked via reflective markers.

%% file: sections/5_Experiment.tex
\section{Simulation Experiments} \label{Sec: Experiments}

We validate and estimate the performance of REBot in the simulation system. In this section, we answer three questions: \textbf{Q1} Can REBot achieve successful evasion under instantaneous DOA? (Sec.~\ref{Subsec: Main results}) \textbf{Q2} What are the robots' reactions under different obstacle conditions? (Sec.~\ref{Subsec: avoidance exp}) \textbf{Q3} How can the rewards' design and recovery stage influence DOA performance? (Sec.~\ref{Subsec: ablation study})

\subsection{Experiment Settings} \label{Subsec: Setup}

\textbf{Tasks}. We conducted simulation experiments in the Isaac gym~\cite{rudin2022learning} to evaluate the DOA ability of the REBot system. 
During testing, the obstacle approaches from diverse directions within a 180° arc in the XZ, YZ, and XY planes of the robot’s body frame (Fig.~\ref{Fig: exp Table}), covering frontal, lateral, overhead, and ground-level threats. The response time is expanded beyond training, with $T_{\text{react}} \in [0.1, 4.0]$~s, allowing evaluation across both immediate reaction and delayed planning scenarios.

\textbf{Metrics}. 
The systems are evaluated with five metrics. The avoidance success rate (\textit{ASR}):  $N_{\text{avoid}} / N_{\text{total}}$; the recovery stability rate (\textit{RSR}):  $N_{\text{recover}} / N_{\text{avoid}}$, indicating the proportion of trials where the robot successfully stabilizes after avoidance; maximum joint power (\textit{MJP}); avoidance moving distance (\textit{AMD}): the base displacement between the robot’s initial and final positions; and gait diversity index (\textit{GDI}): $\mathbb{E}_{s^R \sim \mathcal{D}(s^R)} \left[ \text{Var}_{a^R \sim \pi(a^R|s^R)} [a^R] \right]$ the expected action variance under the learned policy, where $\pi(a^R|s^R)$ denotes the policy distribution over actions at state $s^R$, and $\mathcal{D}(s^R)$ is the state distribution collected during execution.

\textbf{Baselines}.  
1) Agile But Safe (ABS)~\cite{ABS} achieved robust static obstacle avoidance with high-speed navigation motions, without the capability for dynamic obstacles.
2) Reactive RL (RRL)~\cite{Fan_2025} is developed for dynamic obstacle avoidance in the UAV system. The avoidance strategy is based on simplified rigid-body dynamics, which do not generalize to legged whole-body systems.

\begin{figure*}[t]
    \centering
    \vspace{5mm}
    \includegraphics[width=0.85\textwidth]{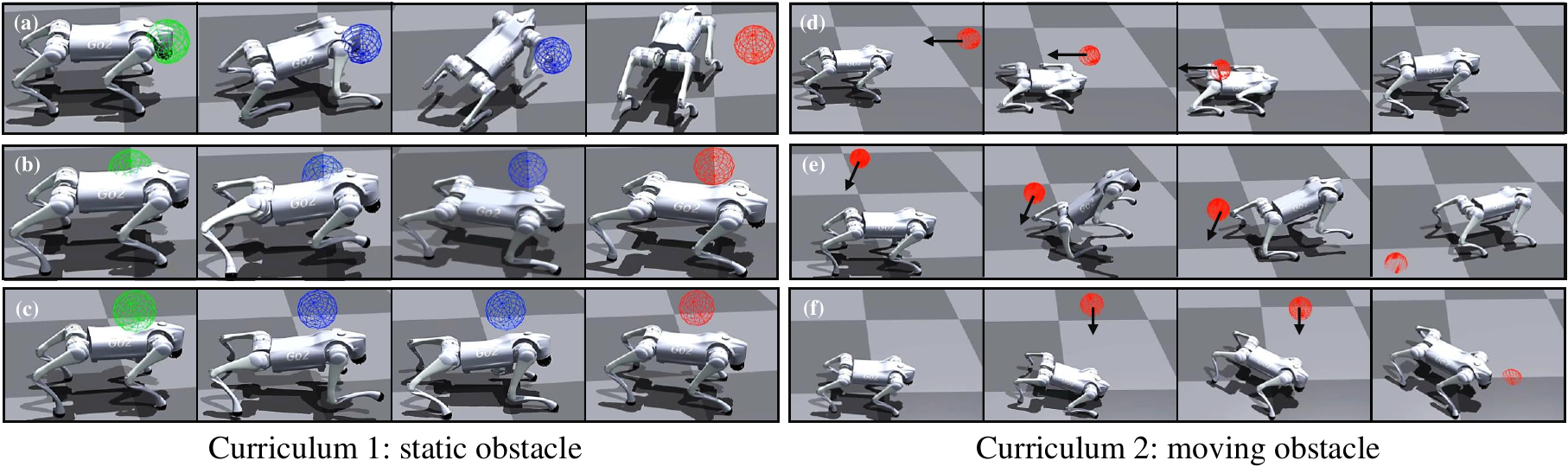}
    \caption{
    Illustration of simulation experiments. \textbf{Curriculum 1}: Static obstacle appears with a delayed time. 
    \textcolor{green}{\CIRCLE} robot in normal stage; 
    \textcolor{blue}{\CIRCLE} robot avoiding the obstacle; 
    \textcolor{red}{\CIRCLE} the obstacle appears and the robot evades the forbidden region. \textbf{Curriculum 2}: The robot avoids the red fast-moving obstacle at all times.}
    \label{Fig: Exp1}
\vspace{-5mm}
\end{figure*}

\begin{table}[t]
    \centering
    \vspace{5mm}
    \renewcommand\arraystretch{1.1}
    \caption{Simulation Experiment Results}
    \label{Table: Simulation results}
    {\scriptsize
    \resizebox{0.87\columnwidth}{!}{
        \begin{threeparttable}
        \begin{tabular}{llccc} 
            \toprule
            $T_{\text{react}}$ / s & Metric & ABS$^\diamond$ & RRL$^\diamond$ & REBot \\
            \midrule
            \multirow{4}{*}{0.1 $\sim$ 0.5} 
                & ASR$\uparrow^*$  & 0.00 & 0.00 & \textbf{0.05} \\
                & RSR$\uparrow^*$  & 0.00 & 0.00 & \textbf{0.03} \\
                & MJP$\downarrow^*$  & 0.51 & 0.52 & \textbf{0.50} \\
                & AMD$\downarrow^*$  & 0.84 & 0.85 & \textbf{0.82} \\
            \midrule
            \multirow{4}{*}{0.5 $\sim$ 1.5} 
                & ASR$\uparrow$  & 0.11 & 0.09 & \textbf{0.65} \\
                & RSR$\uparrow$  & 0.06 & 0.05 & \textbf{0.59} \\
                & MJP$\downarrow$  & 0.52 & 0.51 & \textbf{0.49} \\
                & AMD$\downarrow$  & 0.80 & 0.86 & \textbf{0.47} \\
            \midrule
            \multirow{4}{*}{1.5 $\sim$ 4.0} 
                & ASR$\uparrow$  & 0.51 & 0.41 & \textbf{0.81} \\
                & RSR$\uparrow$  & 0.42 & 0.32 & \textbf{0.74} \\
                & MJP$\downarrow$  & 0.40 & 0.45 & \textbf{0.34} \\
                & AMD$\downarrow$  & 0.60 & 0.70 & \textbf{0.26} \\
            \bottomrule
        \end{tabular}
        \begin{tablenotes}[flushleft]
        \scriptsize
            \item[*] ASR: avoidance success rate; RSR: recovery success rate; MJP: maximum joint power; AMD: avoidance moving distance. \\
            $^\diamond$ ABS: Agile But Safe~\cite{ABS}; RRL: Reactive RL~\cite{Fan_2025}.
        \end{tablenotes}
        \end{threeparttable}
    }}
\vspace{-6mm}
\end{table}

\begin{figure}[htbp]
    \centering
    \includegraphics[width=0.90\columnwidth]{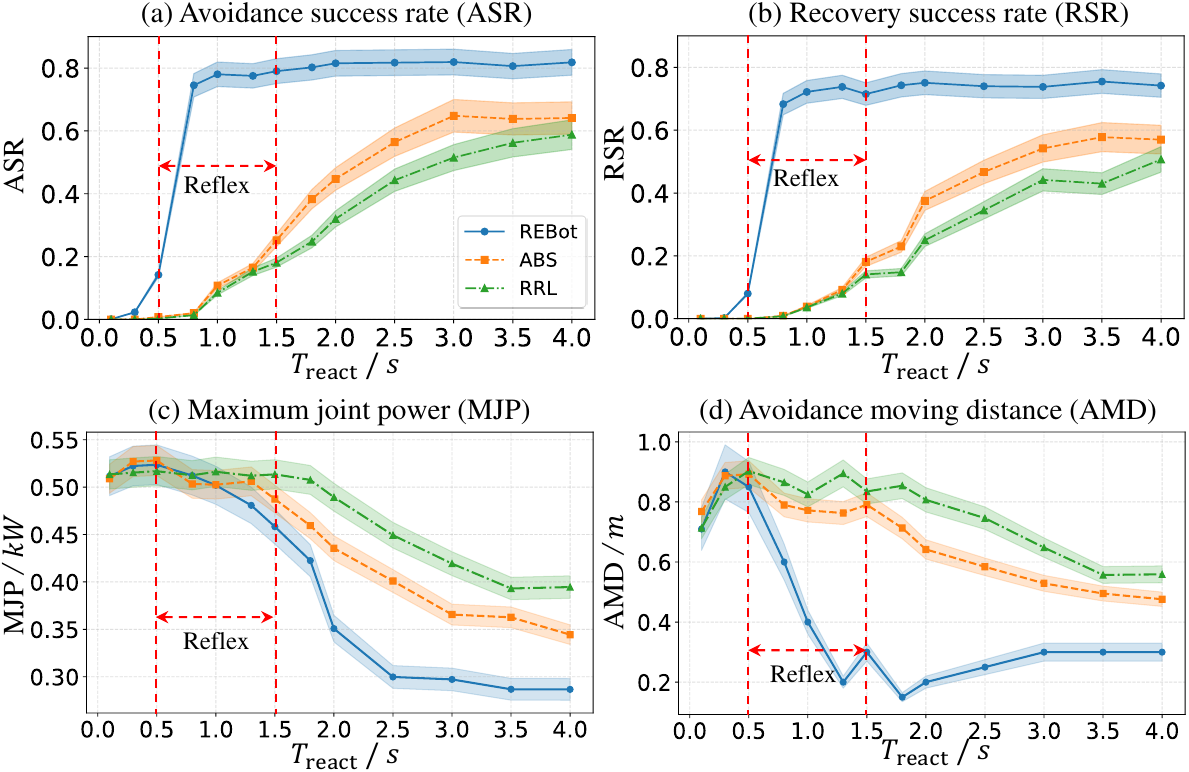}
    \caption{Performances over Reaction Time. Red dashed region indicates reflexive evasion; longer reaction time corresponds to navigation-based avoidance.}
    \label{Fig: success rate over time}
\vspace{-5mm}
\end{figure}

\subsection{Main Experimental Results} \label{Subsec: Main results}

The REBot system is trained in static obstacle and dynamic obstacle avoidance curricula. 
Fig.~\ref{Fig: Exp1} visualizes the simulation experimental results with different obstacle conditions and avoidance behaviors. When obstacles appear in front or on the sides, the robot tends to jump away for evasion (Fig.~\ref{Fig: Exp1}(a)(b)(e)). When obstacles appear on top, the robot will crouch down (Fig.~\ref{Fig: Exp1}(c)(d)(f)). These results demonstrate that REBot enables the robot to select appropriate avoidance strategies based on the obstacle state.

We categorize the reaction time range into three intervals (Tab.~\ref{Table: Simulation results} and Fig.~\ref{Fig: success rate over time}(a)(b)): 0.1 $\sim$ 0.5s, 0.5 $\sim$ 1.5s and 1.5 $\sim$ 4.0s. In the first interval, REBot and both baselines obtain nearly 0 success rates due to insufficient reaction time. In the second interval with moderately short reaction times, REBot exhibits reflexive evasion behaviors, while the two baselines still take low-speed gaits for navigation. This difference leads to much higher ASR and RSR of REBot compared to both baselines. In the third interval,  REBot remains high ASR and RSR. ABS and RRL also improve their performance due to the longer reaction time, but still fall short of REBot because they are not specialized for active DOA. The overall trend, as illustrated in Fig~\ref{Fig: success rate over time}, indicates that REBot effectively addresses instantaneous DOA challenges.

We also explore the relationship between MJP, AMD and reaction time (Fig.~\ref{Fig: success rate over time}(c)(d)). When the reaction time is extremely short (e.g., below 0.5s), we observe a high MJP over 500 W and a large AMD, caused by intense evasion behaviors such as jumping away. In the moderately short reaction time interval, both MJP and AMD decrease, where REBot can adopt more appropriate avoidance behaviors such as crouching down. With longer reaction time, both MJP and AMD converge to lower values, as REBot has enough time to execute smoother navigation-based avoidance behaviors. The trends show that REBot adapts avoidance behaviors based on time constraints and avoidance efficiency.

\subsection{Analysis of Avoidance Ability} \label{Subsec: avoidance exp}

\begin{figure*}[t]
    \centering
    \vspace{5mm}
    \includegraphics[width=0.85\textwidth]{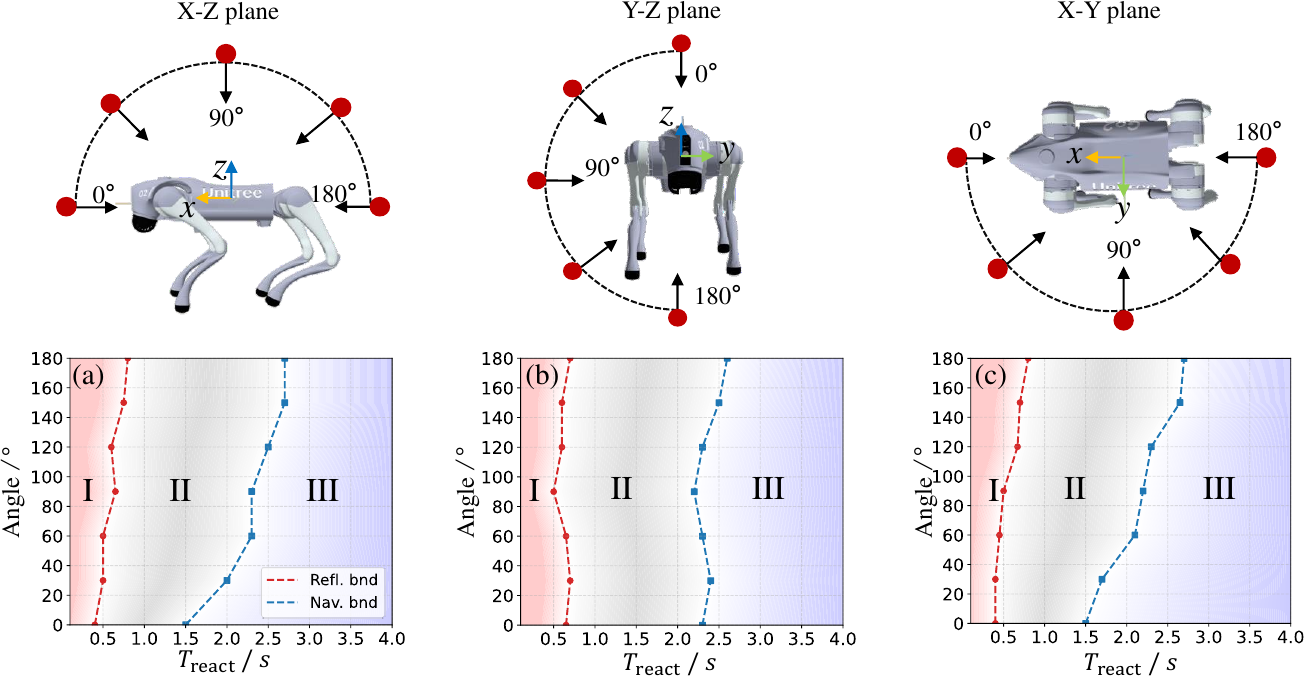}
    \caption{Robot avoids obstacles in various directions and reaction time. \textbf{Top row} shows the obstacles' approaching directions in three planes. \textbf{Bottom row} figures indicate the different avoidance behaviors. Region I: avoidance failure, II: reflexive evasion, III: navigation avoidance.}
    \label{Fig: exp Table}
\vspace{-4mm}
\end{figure*}

We evaluate the effect of the obstacle direction on the robot's avoidance ability by applying impacts from different angles within the X-Z, Y-Z and X-Y planes of the robot's body frame (Fig.~\ref{Fig: exp Table}). Based on ASR and MJD, we divide the robot's avoidance behavior space into three regions: region I, where the robot fails to avoid; region II, where the robot adopts reflexive evasion; and region III, where the robot adopts navigation-based avoidance. The boundary between region I and II is defined by ASR over 30\%, while the boundary between region II and III is defined by MJD below 300 W. 

In the X-Z and X-Y planes (Fig.~\ref{Fig: exp Table}(a)(c)), we observe that obstacles appearing in front of the robot are easier to avoid, and navigation-based avoidance can be achieved with shorter reaction time. In contrast, obstacles approaching from the back require longer reaction time and make navigation-based avoidance more difficult. This asymmetry is attributed to the mechanical design of Unitree Go2, where the leg structure facilitates faster backward motion but makes forward jumping more challenging. In the Y-Z plane (Fig.~\ref{Fig: exp Table}(b)), we find that obstacles approaching from the sides are easier to avoid compared to those from the top or bottom, with a shorter transition time from reflexive evasion to navigation-based avoidance. These results highlight the robot's avoidance capability varies significantly depending on the obstacle direction.

\subsection{Ablation Studies of REBot System} \label{Subsec: ablation study}

\begin{table}[t]
    \centering
    \renewcommand\arraystretch{1.1}
    \caption{Ablation Performances of REBot}
    \label{Table: task results}
    {\scriptsize
    \resizebox{0.95\columnwidth}{!}{
        \begin{threeparttable}
        \begin{tabular}{llcccc}
            \toprule
            $T_\text{react}$ / s & Metric & w/o rcv.$^1$ & w/o curr.$^2$ & w/o adp.$^3$ & REBot \\
            \midrule
            \multirow{3}{*}{0.5$\sim$1.5} 
                & ASR$\uparrow^*$   & 0.63 & 0.48 & 0.59 & \textbf{0.65} \\
                & RSR$\uparrow^*$  & 0.31 & 0.39 & 0.51 & \textbf{0.59} \\
                & GDI$\uparrow^*$  & 2.46 & 2.41 & 1.43 & \textbf{2.51} \\
            \midrule
            \multirow{3}{*}{1.5$\sim$4.0} 
                & ASR  & 0.80 & 0.71 & 0.78 & \textbf{0.81} \\
                & RSR  & 0.63 & 0.60 & 0.69 & \textbf{0.74} \\
                & GDI  & 2.06 & 2.24 & 1.36 & \textbf{2.13} \\
            \bottomrule
        \end{tabular}
        \begin{tablenotes}[flushleft]
        \begin{small}
            \item[*] ASR: avoidance success rate; RSR: recovery success rate; GDI: gait diversity index. \\
            $^1$ w/o recovery stage; $^2$ w/o curriculum one learning; $^3$ w/o adaptive reward in avoidance policy.
        \end{small}
        \end{tablenotes}
        \end{threeparttable}
    }}
\vspace{-2mm}
\end{table}

\begin{figure}[t]
    \centering
    \includegraphics[width=0.95\columnwidth]{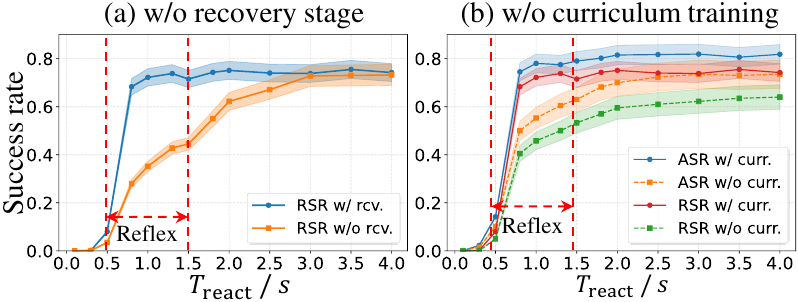}
    \caption{Success Rate of Ablation Studies. Red dashed region indicates reflexive evasion; longer reaction time corresponds to navigation-based avoidance.}
    \label{Fig: task steps}
\vspace{-5mm}
\end{figure}

\textbf{The recovery stage ensures a stable standing posture after rapid reflexive evasion. }
To validate its effectiveness, we compare the performance of the REBot system with and without the recovery policy. 
As shown in Table~\ref{Table: task results} and Fig.~\ref{Fig: task steps}(a), removing the recovery stage leads to a drop (20\%) in the success rate within the reflex region. 
Additionally, as reaction time increases, avoidance behaviors shift from reflex to navigation, reducing the influence of the recovery stage on robot stabilization.

\textbf{Curriculum learning enables a smooth transition from normal stage to fast-moving reflexive evasion.}
In the first stage, the robot learns to avoid obstacles that appear suddenly at varying positions; in the second stage, it generalizes to obstacles approaching from different directions.  
An ablation study reveals the importance of this progressive training: when policies are trained directly on the second curriculum (bypassing the first), Table~\ref{Table: task results} and Fig.~\ref{Fig: task steps}(b) show removing the first curriculum causes a 5\% decline in both ASR and RSR for both reflexive evasion and navigation avoidance. This performance gap stems from the more moderate and diverse gaits learned during static obstacle avoidance, which prove beneficial for a stabilized start when facing fast-moving obstacles.

\textbf{The adaptive reward encourages diversified avoidance gaits that improve avoidance robustness.} Without this term, the reinforcement learning algorithm converges to a single avoidance behavior (e.g., consistently jumping backward) regardless of obstacle variations. We conduct an ablation study to evaluate the adaptive reward's effectiveness. As shown in Tab.~\ref{Table: task results}, removing the adaptive reward results in: (1) a significant 40\% decrease in gait diversity index (GDI), demonstrating reduced behavioral diversity, and (2) moderate declines in both ASR and RSR. These findings indicate that the gait diversity promoted by the adaptive reward contributes directly to improved avoidance performance.

%% file: sections/6_real_robot.tex
\section{Real-Robot Demonstration} \label{Sec: real robot demo}

\begin{figure*}[t]
    \centering
    \vspace{5mm}
    \includegraphics[width=0.85\textwidth]{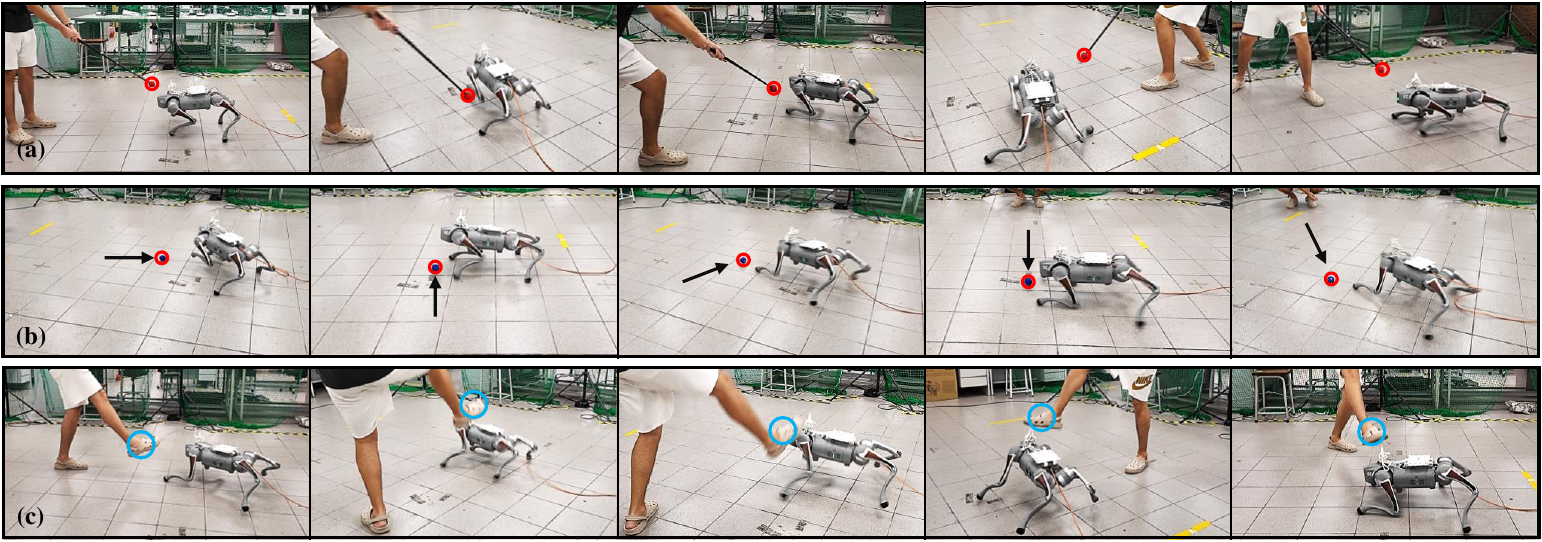}
    \caption{REBot system real-robot demonstrations on Unitree Go2 Robot (See \href{https://rebot-2025.github.io/}{video}). (a) the robot is poked from different directions using a stick; (b) a ball is launched towards the robot from different directions; (c) the robot is subjected to intentional kicks from different directions.}
    \label{Fig: exp real}
\vspace{-5mm}
\end{figure*}

We deploy the REBot system on a Unitree Go2 robot to demonstrate its reflexive evasion capabilities in real-world scenarios. The robot and obstacles are tracked using an OptiTrack motion capture system to provide accurate position information. To generate diverse dynamic obstacles, we test three interaction types: poking with a stick (Fig.~\ref{Fig: exp real}(a)), throwing a ball (Fig.~\ref{Fig: exp real}(b)), and kicking (Fig.~\ref{Fig: exp real}(c)), each targeting the robot from different directions, including front, left, right, left-front, and right-front. 

When an obstacle approaches, the robot triggers the avoidance mode to perform reflexive evasion. The primary avoidance actions include jumping away from the obstacle and crouching down. After completing the avoidance maneuver, the robot switches to the recovery mode to regain a stable standing posture. Additionally, we observe that when the poking motion is relatively slow, the robot tends to adopt navigation-based avoidance strategies instead of reflexive actions, leveraging the longer available reaction time to perform smoother behaviors.Under the real-world test conditions, the REBot system achieves an ASR of 56\% and an RSR of 53\%. The performance gap compared to simulation is mainly attributed to Sim2Real challenges such as unmodeled actuator dynamics, latency in control execution, and surface friction variability, which particularly affect fast reflexive responses requiring precise torque delivery.

%% file: sections/7_conclusion.tex
\section{Conclusion} \label{Sec: Conclusion}

We initiate the study of reflexive evasion as a critical capability for dynamic obstacle avoidance in quadrupedal robots, where traditional navigation-based methods fall short under tight reaction constraints. To address this challenge, we develop REBot, a unified control system that couples rapid avoidance and stability recovery through reinforcement learning and structured training strategies. Extensive experiments in simulation and on real hardware confirm REBot's ability to perform reliable, adaptive evasive maneuvers, while revealing important characteristics of reflexive responses shaped by robot morphology and obstacle dynamics. Our results point toward new directions for building more agile, resilient, and safety-aware legged robotic systems in dynamic environments.

Despite its effectiveness, REBot has several limitations. First, while we focus on the reflexive evasion policy, we leave precise obstacle position perception as an assumption; independent perception will be addressed in next work. Second, the avoidance behavior is biased by the hardware design of the robot, showing a preference for backward jumps due to leg morphology. Lastly, a Sim2Real gap remains in low-level motor control, as joint velocity commands in simulation do not fully capture the dynamics of torque-based servo actuation in high-speed evasive maneuvers, nor the variability introduced by real-world ground friction.

%% file: sections/11_appendix.tex
\section{RL Policy Training Details}

We trained two separate RL policies using PPO: an avoidance policy focused on dynamic obstacle evasion and a recovery policy responsible for post-disturbance stabilization. Both policies used actor-critic architectures implemented as multilayer perceptrons (MLPs) with hidden layers of 512, 256, and 128 units, using ELU activations.

For optimization, we applied a clipped surrogate objective with a clip parameter $\epsilon = 0.2$ (Tab.~\ref{TAB:PPO hyperparameter}), an entropy coefficient $c_2 = 0.01$, and a value loss coefficient $c_1 = 1.0$. We used the Adam optimizer with an initial learning rate $\alpha = 1 \times 10^{-3}$, combined with adaptive learning rate scheduling. The discount factor was set to $\gamma = 0.99$, and the generalized advantage estimation (GAE) parameter was set to $\lambda = 0.95$. Each PPO iteration collected 24 steps per environment across 4096 parallel environments, followed by 5 learning epochs over 4 mini-batches. The maximum gradient norm was clipped at 1.0, and the desired KL divergence threshold was set to 0.01. Training was conducted using Isaac Gym on an NVIDIA RTX 4090 GPU.

\begin{table}[H]
    \centering
    \caption{Summary of PPO hyperparameters used for training}
    \label{TAB:PPO hyperparameter}
    \resizebox{0.75\columnwidth}{!}{  
        \begin{tabular}{lc}
            \toprule
            Hyperparameter & Value \\
            \midrule
            Actor hidden layers         & [512, 256, 128] \\
            Critic hidden layers        & [512, 256, 128] \\
            Activation                 & ELU \\
            Learning rate ($\alpha$)              & $1 \times 10^{-3}$ \\
            Clip parameter ($\epsilon$)             & 0.2 \\
            Value loss coefficient ($c_1$)     & 1.0 \\
            Entropy coefficient ($c_2$)        & 0.01 \\
            Discount factor ($\gamma$) & 0.99 \\
            GAE parameter ($\lambda$)  & 0.95 \\
            Desired KL divergence      & 0.01 \\
            Max gradient norm          & 1.0 \\
            Steps per env per iter     & 24 \\
            Mini-batches per iter      & 4 \\
            Learning epochs per iter   & 5 \\
            Max iterations             & 5000 \\
            Parallel environments      & 4096 \\
            \bottomrule
        \end{tabular}
    }
\end{table}

Building on the PPO optimization framework, the avoidance policy is guided by a reward structure that combines three main components: avoidance rewards that encourage maintaining safe distances from dynamic obstacles and penalize collisions, regularization rewards that promote stable, symmetric, and energy-efficient gait patterns, and adaptive rewards that foster motion diversity, speed adaptation, and directional efficiency under varying threat levels. This combination ensures that the robot can execute timely evasive maneuvers while maintaining locomotion stability and natural gait coordination.

In parallel, the recovery policy employs a dedicated reward design focused on regaining upright posture, minimizing joint velocities, returning to the nominal base position, and ensuring smooth, low-torque recovery transitions after disturbances. This structure enables the robot to rapidly restore balance and seamlessly transition back to its default locomotion behaviors after a disturbance or evasive event.

In addition to these task-specific rewards, we incorporate a set of auxiliary regularization terms to enforce physical plausibility, smoothness, and mechanical safety. These terms play a critical role in constraining the robot’s low-level dynamics, preventing unrealistic or unsafe behaviors, and improving the overall robustness and hardware transferability of the learned policies. The complete set of these auxiliary terms is summarized in Tab.~\ref{TAB:Auxiliary regularization terms}.

\begin{table}[H]
    \centering
    \caption{Summary of auxiliary regularization terms}
    \label{TAB:Auxiliary regularization terms}
    \resizebox{\columnwidth}{!}{ 
        \begin{tabular}{cl}
            \toprule
            Term & Purpose \\
            \midrule
            $|v^{R,z}_t|^2$ & Penalize vertical velocity \\
            $\|\dot{\theta}^{R,xy}_t\|$ & Penalize horizontal angular velocity \\
            $\|\theta^{R,xy}_t\|$ & Penalize non-flat orientation \\
            $\sum_i (a^{R,i}_t - a^{R,i}_{t-1})^2$ & Penalize abrupt action changes \\
            $\sum_i \mathbf{1}_c^i$ & Penalize body collisions \\
            $||v^{R,xy}_t - v^{R,xy,\text{cmd}}_t||$ & Track command linear velocity \\
            $|\omega^{R,z}_t - \omega^{R,z,\text{cmd}}_t|$ & Track command angular velocity \\
            $\sum_{i} \left( t^i_{\text{air}} - 0.5 \right)$ & Reward long foot swing phases \\
            $\mathbf{1} \left( \max_{i} \left( \left\lVert \mathbf{f}^{R,xy,i}_t \right\rVert / \left| f^{R,z,i}_t \right| \right) > 5 \right)$ & Penalize stumbling events \\
            $\sum_i \|f^{R,i}_t - f^{R,i}_{\text{th}}\|^2$ & Penalize excessive foot contact forces \\
            \bottomrule
        \end{tabular}
    }
\end{table}

\section{Domain Randomization \& Curricula}

To enhance policy generalization and robustness, we applied domain randomization and a staged curriculum strategy during training. Domain randomization (Tab.~\ref{TAB:domainrand}) introduces variability across observation parameters (e.g., joint position and velocity noise), dynamics parameters (e.g., ground friction, added base mass, and obstacle velocity), and episode-level parameters (e.g., commanded yaw and episode duration). By sampling these parameters uniformly within predefined ranges at the start of each episode, the policy is exposed to a diverse set of conditions, improving its ability to handle modeling uncertainties, mitigate overfitting to narrow simulation settings, and transfer reliably to real-world deployment.

\begin{table}[h]
    \centering
    \caption{Domain Randomization Settings for Policy Training}
    \label{TAB:domainrand}
    \resizebox{0.77\columnwidth}{!}{
    \begin{tabular}{ll} 
        \toprule
        Term & Value\\ 
        \midrule
        \textbf{Observation} & \\
        \quad Joint position noise & $\mathcal{U}(-0.01,0.01)$~rad\\
        \quad Joint velocity noise & $\mathcal{U}(-1.5,1.5)$~rad/s\\
        \quad Angular velocity noise & $\mathcal{U}(-0.2,0.2)$~rad/s\\
        \quad Projected gravity noise & $\mathcal{U}(-0.05,0.05)$~m/s\textsuperscript{2}\\ 
        \quad Height measure noise & $\mathcal{U}(-0.1,0.1)$~m\\
        \textbf{Dynamics} & \\
        \quad Friction factor & $\mathcal{U}(0.5,1.25)$\\ 
        \quad Added base mass & $\mathcal{U}(-1.0,1.0)$~kg\\ 
        \quad Obstacle position (per axis) & $\mathcal{U}(-0.4,0.4)$~m\\
        \quad Obstacle radius & $\mathcal{U}(0.05,0.3)$~m\\
        \quad Obstacle velocity & $\mathcal{U}(1.0,6.0)$~m/s\\
        \quad Reaction time & $\mathcal{U}(0.1,4.0)$~s\\ 
        \textbf{Episode} &  \\
        \quad Episode length & $\mathcal{U}(8.0,10.0)$~s\\ 
        \quad Command robot yaw & $\mathcal{U}(-1.0,1.0)$~rad\\ 
        \quad Command robot velocity & $\mathcal{U}(-1.0,1.0)$~m/s\\
        \quad Command robot heading & $\mathcal{U}(-\pi,\pi)$~rad\\
        \bottomrule
    \end{tabular}
    }
    \vspace{-3mm}
\end{table}

\begin{figure*}[t]
    \centering
    \vspace{5mm}
    \includegraphics[width=0.85\textwidth]{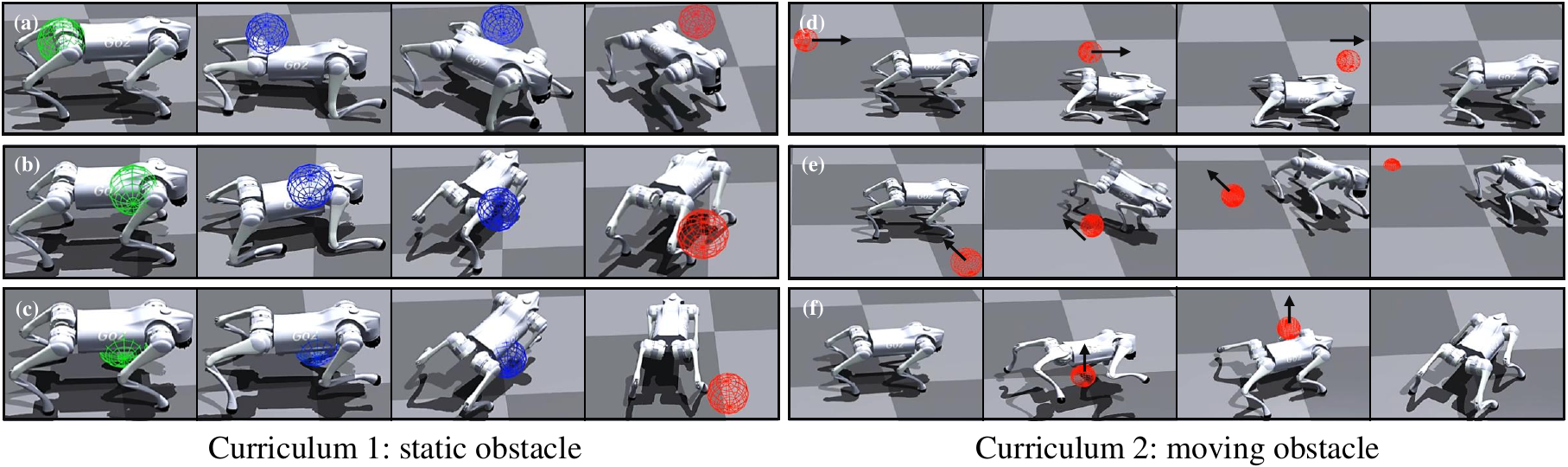}
    \caption{Additional results of simulation experiments. (a) and (d) show the obstacle hits from the back; (b) and (e) hit from the right; (c) and (f) hit from the bottom.}
    \label{Fig: App Exp sim}
\vspace{-4mm}
\end{figure*}

The curriculum learning strategy is implemented to gradually increase task complexity. Initially, the policy learns to avoid static obstacles that suddenly appear at specific positions near the robot, without any external perturbations. In the next stage, dynamic obstacles with varying speeds and trajectories are introduced, requiring the robot to perform rapid, adaptive avoidance maneuvers. Finally, disturbances and environmental uncertainties are incorporated to ensure the policy remains stable and robust under real-world deployment conditions. Additional experimental results are provided in Fig.~\ref{Fig: App Exp sim}.

\section{Real-Robot Experiment Settings}

\begin{figure}[t]
    \centering
    \includegraphics[width=0.9\columnwidth]{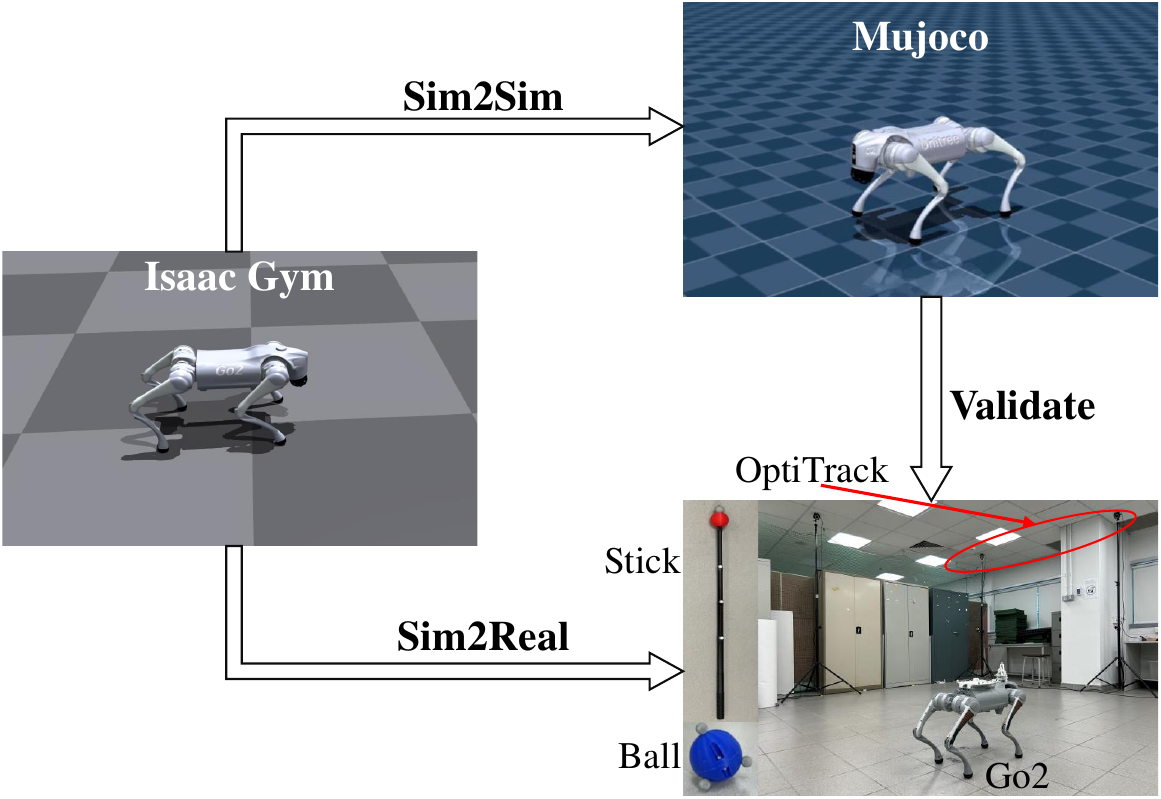}
    \caption{Transfer pathways from Isaac Gym to MuJoCo (sim2sim) and to the real robot (sim2real). Real-robot experiments involve dynamic obstacles including a ball, a stick, and a human foot, representing diverse scenarios.}
    \label{Fig: sim2sim2real}
\vspace{-4mm}
\end{figure}

To evaluate the real-world feasibility of the proposed avoidance-recovery strategy, experiments were conducted on a Unitree Go2 quadrupedal robot. Prior to deployment, the learned policies were validated through sim-to-sim transfer from Isaac Gym to MuJoCo to ensure robustness under a higher-fidelity simulation environment (Fig.\ref{Fig: sim2sim2real}). Only after passing these intermediate robustness tests were the policies transferred to the real robot.

For the real-robot setup, we employed an OptiTrack motion capture system to provide precise localization of both the robot and the dynamic obstacles (Fig.~\ref{Fig: sim2sim2real}). The system offers sub-millimeter positioning accuracy, enabling high-fidelity state tracking without relying on onboard perception (e.g., cameras or LiDAR). This design ensures that the robot directly receives ground truth position and velocity information for both itself and the obstacles at each control step.

To simulate various dynamic threats, we used three types of physical obstacles: a rigid ball, a stick with a marker-attached tip, and a human foot. Marker points were affixed to each obstacle, allowing their motion to be tracked and fed into the robot’s control pipeline. During experiments, the Go2 executed the trained avoidance and recovery policies in real time, responding to incoming obstacles by performing reflexive evasion and subsequent stabilization maneuvers.

\section{Additional Experiment Results}

\begin{figure*}[t]
    \centering
    \includegraphics[width=0.85\textwidth]{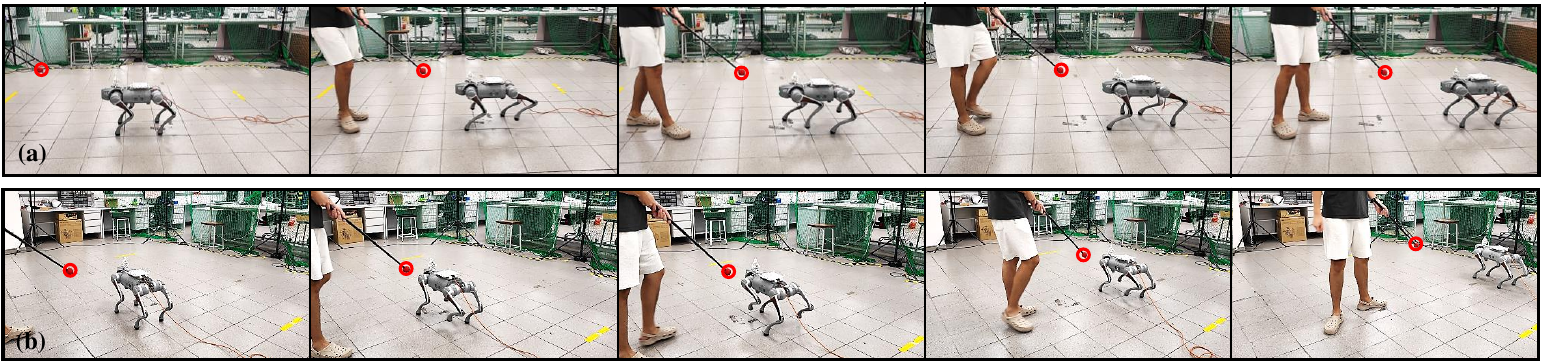}
    \caption{Navigation-based avoidance under slow-moving stick disturbances, providing sufficient reaction time. (a) Poking from the front; (b) poking from the left.}
    \label{Fig: App Exp real Navi}
\vspace{-2mm}
\end{figure*}

We present additional qualitative results to illustrate the effectiveness and versatility of the proposed avoidance-recovery strategy across diverse scenarios. Figure~\ref{Fig: App Exp real Navi} shows an example where the robot employs navigation-based avoidance strategies, relying on trajectory adjustment rather than reflexive maneuvers. This is feasible because the approaching obstacles are slow, providing sufficient reaction time for planned avoidance.

\begin{figure*}[t]
    \centering
    \includegraphics[width=0.85\textwidth]{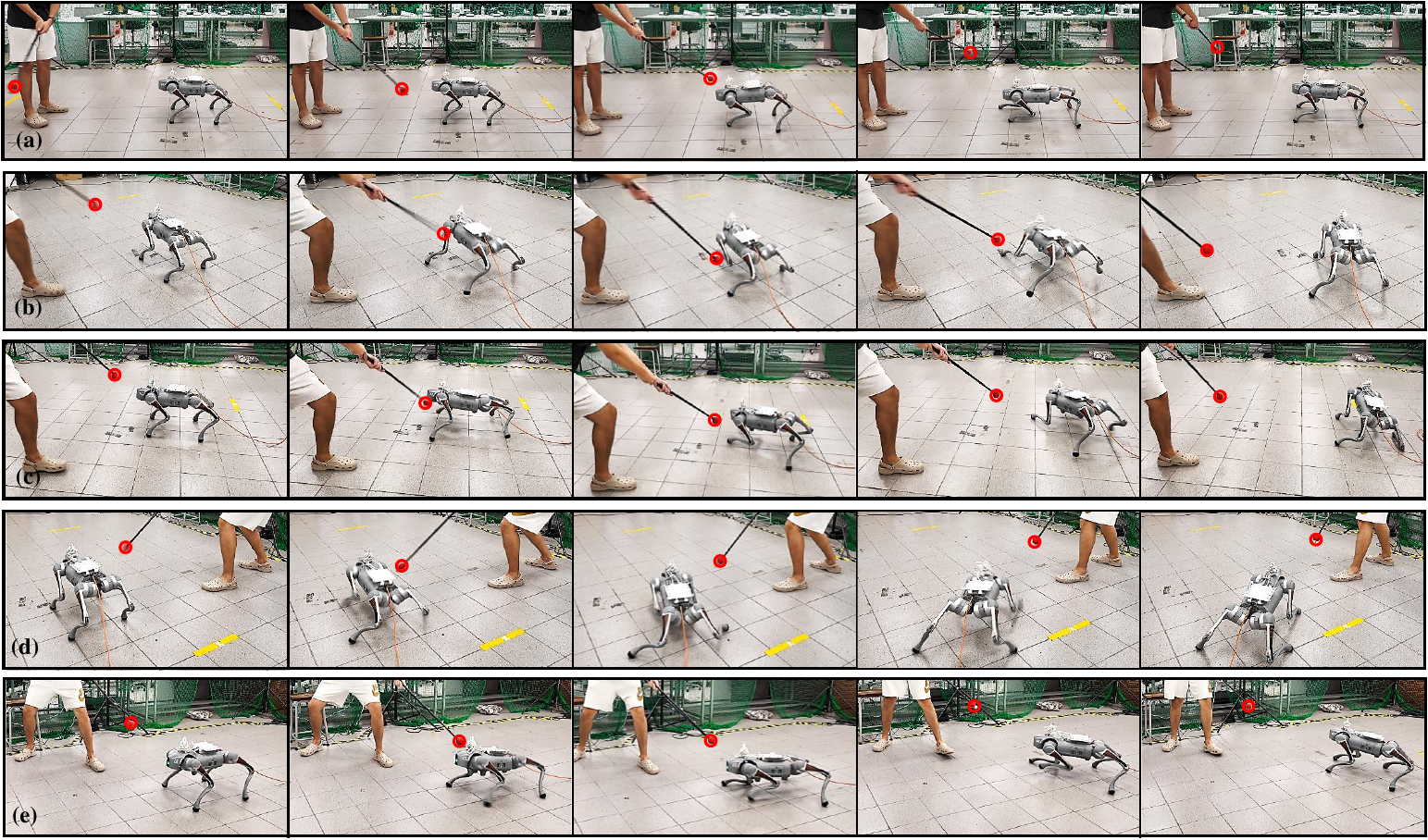}
    \caption{Reflexive evasion under fast stick disturbances with short reaction time. (a) From the front; (b) from the left; (c) from the left front; (d) from the right; (e) from the right front.}
    \label{Fig: App Exp real Stick}
\vspace{-2mm}
\end{figure*}

Figure~\ref{Fig: App Exp real Stick} demonstrates the robot’s reflexive response under stick-induced prodding attacks from various directions, showcasing its ability to rapidly adjust posture and evade external physical disturbances.

\begin{figure*}[t]
    \centering
    \includegraphics[width=0.85\textwidth]{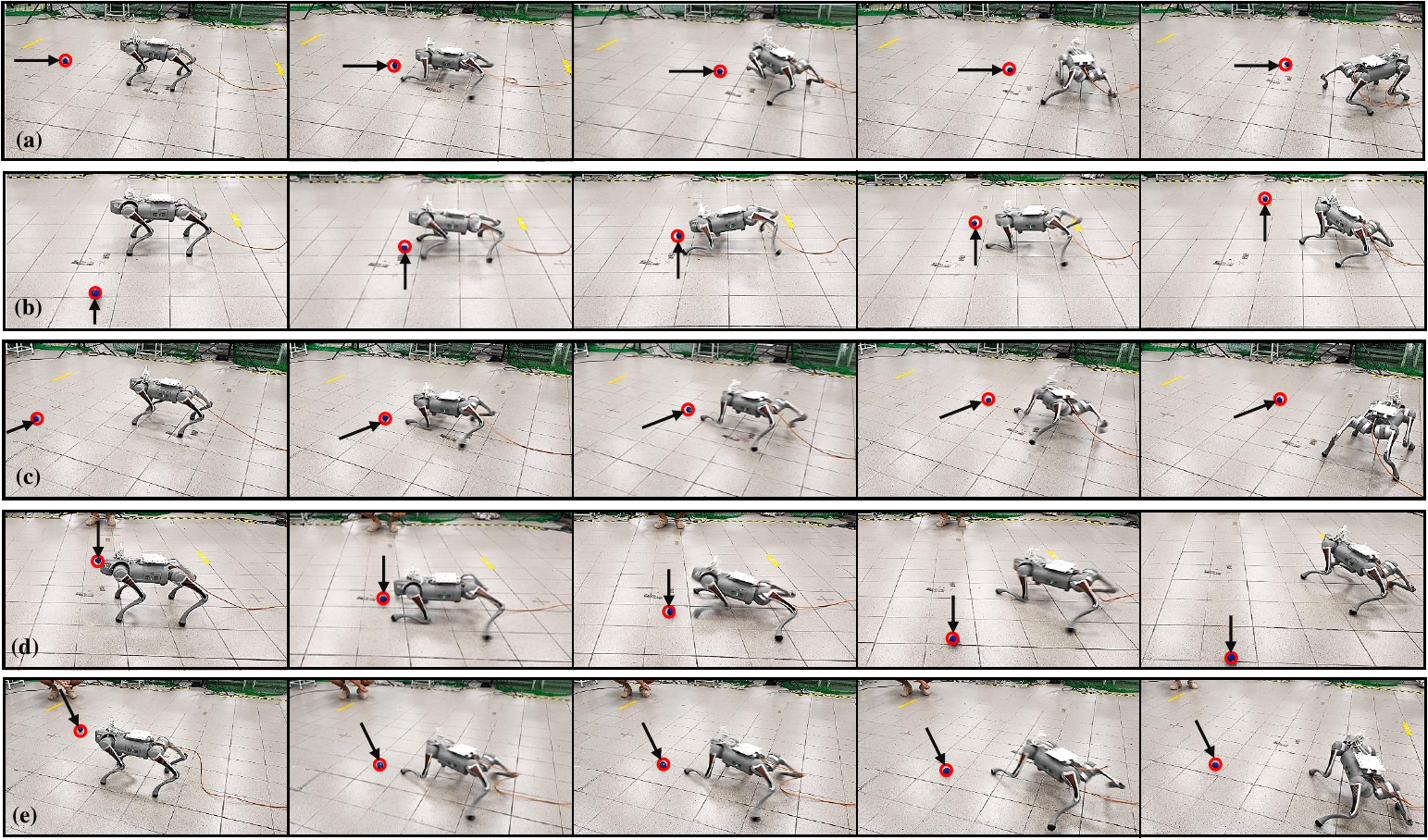}
    \caption{Reflexive evasion under ball-throw impacts from different directions. (a) From the front; (b) from the left; (c) from the left front; (d) from the right; (e) from the right front.}
    \label{Fig: App Exp real Ball}
\vspace{-2mm}
\end{figure*}

Figure~\ref{Fig: App Exp real Ball} presents results where a ball is thrown toward the robot from multiple angles, testing the policy’s capacity to execute fast evasive maneuvers under short reaction times.

\begin{figure*}[t]
    \centering
    \includegraphics[width=0.85\textwidth]{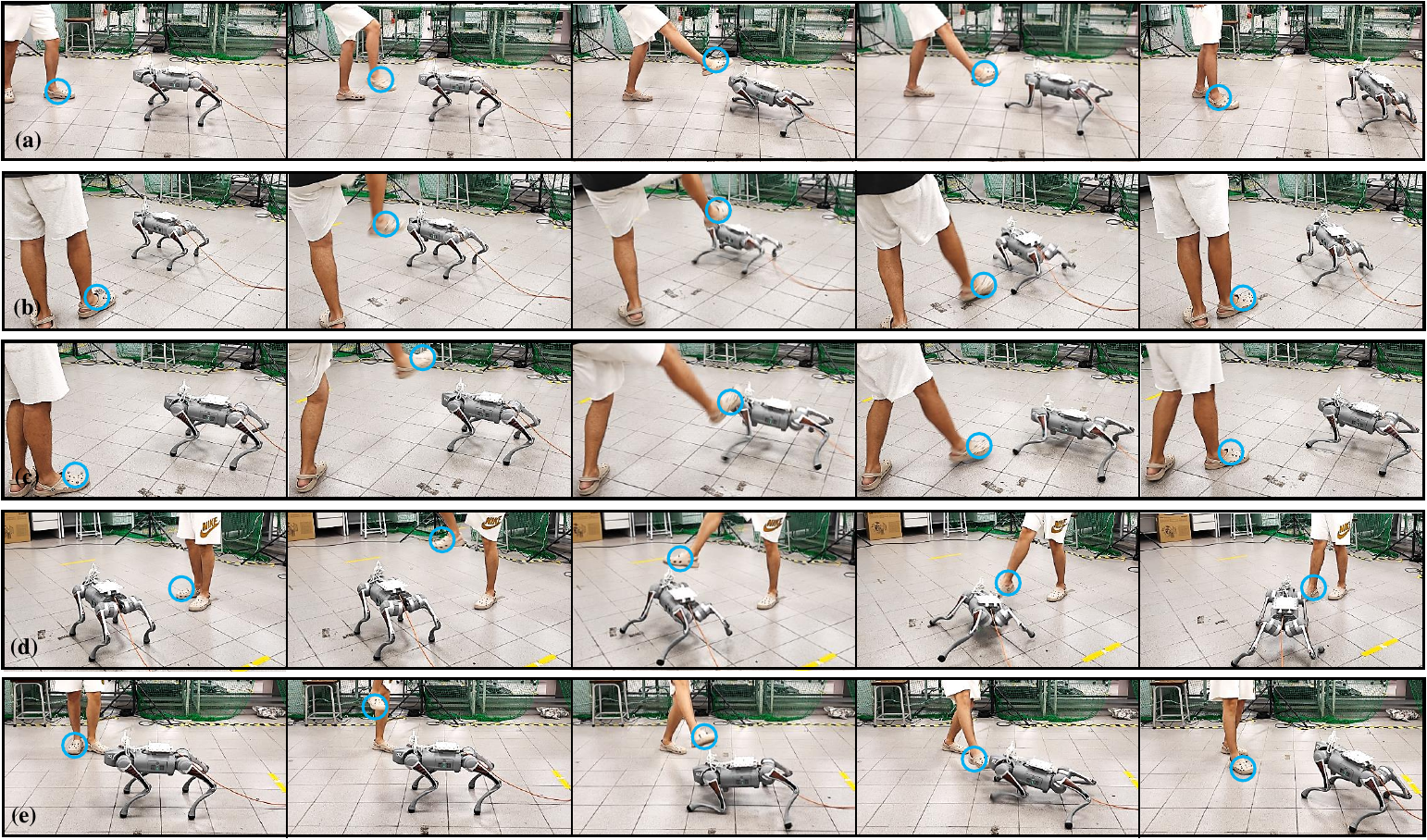}
    \caption{Reflexive evasion under foot-kick disturbances from different directions. (a) From the front; (b) from the left; (c) from the left front; (d) from the right; (e) from the right front.}
    \label{Fig: App Exp real Foot}
\vspace{-4mm}
\end{figure*}

Finally, Figure~\ref{Fig: App Exp real Foot} highlights experiments where the robot faces unexpected kicks from a human foot at different approach angles, demonstrating the policy’s robustness in handling unstructured, real-world disturbances.

\clearpage